%% file: savlm_icra25.tex
\documentclass[letterpaper, 10 pt, conference]{ieeeconf}  % Comment this line out if you need a4paper

\IEEEoverridecommandlockouts                              % This command is only needed if 
\makeatletter
\let\NAT@parse\undefined
\makeatother

\usepackage{graphicx} % for pdf, bitmapped graphics files
\usepackage{kotex}
\usepackage{xcolor}
\usepackage{listings}
\usepackage{amsmath} 
\usepackage{multirow}
\usepackage{booktabs}
\usepackage{hhline}
\usepackage{listings}
\usepackage{tcolorbox}
\usepackage{hyperref}

\title{\LARGE \bf

Space-Aware Instruction Tuning: Dataset and Benchmark \\for Guide Dog Robots Assisting the Visually Impaired
}
\author{ByungOk Han*$^{1}$, Woo-han Yun*$^{1}$, Beom-Su Seo$^{1}$, and Jaehong Kim$^{1}$% <-this % stops a space
\thanks{ByungOk Han* and Woo-han Yun* contributed equally to this work and are co-corresponding authors.\{byungok.han, yochin\}@etri.re.kr}
\thanks{$^{1}$ETRI, Daejeon, Republic of Korea}
\thanks{% GUIDE DOG
This work was supported by the Institute of Information \& communications Technology Planning \& Evaluation(IITP) grant funded by the Korea government(MSIT) (RS-2023-00215760, Guide Dog: Development of Navigation AI Technology of a Guidance Robot for the Visually Impaired Person).
This research (paper) used datasets from `The Open AI Dataset Project (AI-Hub, S. Korea)'. All data information can be accessed through `AI-Hub (www.aihub.or.kr)'.}% <-this % stops a space
}

\begin{document}

\maketitle
\thispagestyle{empty}
\pagestyle{empty}

%%%%%%%%%%%%%%%%%%%%%%%%%%%%%%%%%%%%%%%%%%%%%%%%%%%%%%%%%%%%%%%%%%%%%%%%%%%%%%%%
\begin{abstract}
\input{tex/abstract}
\end{abstract}

%%%%%%%%%%%%%%%%%%%%%%%%%%%%%%%%%%%%%%%%%%%%%%%%%%%%%%%%%%%%%%%%%%%%%%%%%%%%%%%%
\input{tex/intro}

%%%%%%%% 제목에 따라서 다르게 준비
%\input{tex/related_work}
%%%%%%%% VLM, 벤치마크/데이터셋 등
\input{tex/method.tex}
\input{tex/results}
%%%%%%%% 실험은 2가지 종류만?? num word까지 세종류? -
% \input{tex/limitation}
% \input{tex/limitation}
\input{tex/conclusion}

%%%%%%%% 한계점??

\input{savlm_icra25.bbl}
% \clearpage
% \input{tex/appendix}

\end{document}

%% file: tex/abstract.tex
Guide dog robots offer promising solutions to enhance mobility and safety for visually impaired individuals, addressing the limitations of traditional guide dogs, particularly in perceptual intelligence and communication. With the emergence of Vision-Language Models (VLMs), robots are now capable of generating natural language descriptions of their surroundings, aiding in safer decision-making. However, existing VLMs often struggle to accurately interpret and convey spatial relationships, which is crucial for navigation in complex environments such as street crossings. We introduce the Space-Aware Instruction Tuning (SAIT) dataset and the Space-Aware Benchmark (SA-Bench) to address the limitations of current VLMs in understanding physical environments. Our automated data generation pipeline focuses on the virtual path to the destination in 3D space and the surroundings, enhancing environmental comprehension and enabling VLMs to provide more accurate guidance to visually impaired individuals. We also propose an evaluation protocol to assess VLM effectiveness in delivering walking guidance. Comparative experiments demonstrate that our space-aware instruction-tuned model outperforms state-of-the-art algorithms. We have fully open-sourced the SAIT dataset and SA-Bench, along with the related code, at \href{https://github.com/byungokhan/Space-awareVLM}{https://github.com/byungokhan/Space-awareVLM}.

%% file: tex/intro.tex
\section{Introduction}
Guide dog robots have the potential to significantly improve mobility and safety for the visually impaired, with continued research driving their advancement \cite{wei2014guide}\cite{xiao2021robotic}\cite{hong2024collaborative}\cite{saegusa2010development}.
Unlike living guide dogs, robots are expected to overcome limitations in perceptual intelligence, such as a restricted ability to understand their surroundings and the one-way nature of communication \cite{Hochul2024}. 
Recent advancements in Vision-Language Models (VLMs), which enable natural language scene descriptions, allow robots to more effectively convey critical situational information \cite{chen2020uniter}\cite{alayrac2022flamingo}\cite{you2023ferret}\cite{li2023blip}.
To enhance safe decision-making for blind and low-vision individuals at complex street crossings, the use of VLMs is explored to interpret scenes and generate safety scores and descriptions in natural language \cite{hochul2024crossstreet}.
As a dataset focused on walking environments for the visually impaired, the SideGuide Dataset \cite{Kibaek2020} offers images from these scenarios, along with instance-level annotations for object detection and segmentation, and dense disparity maps.
Visually Impaired Assistance with Large Models \cite{zhao2024vialmsurveybenchmarkvisually} defines a task for providing step-by-step guidance to the visually impaired using VLMs and proposes a benchmark for evaluating various state-of-the-art models. 
However, the benchmark is restricted to environments such as homes and supermarkets, which are not as critical as walking environments for the visually impaired.

\begin{figure}[t]
    \centering
    % \vspace{-3pt}
    \includegraphics[width=\linewidth]{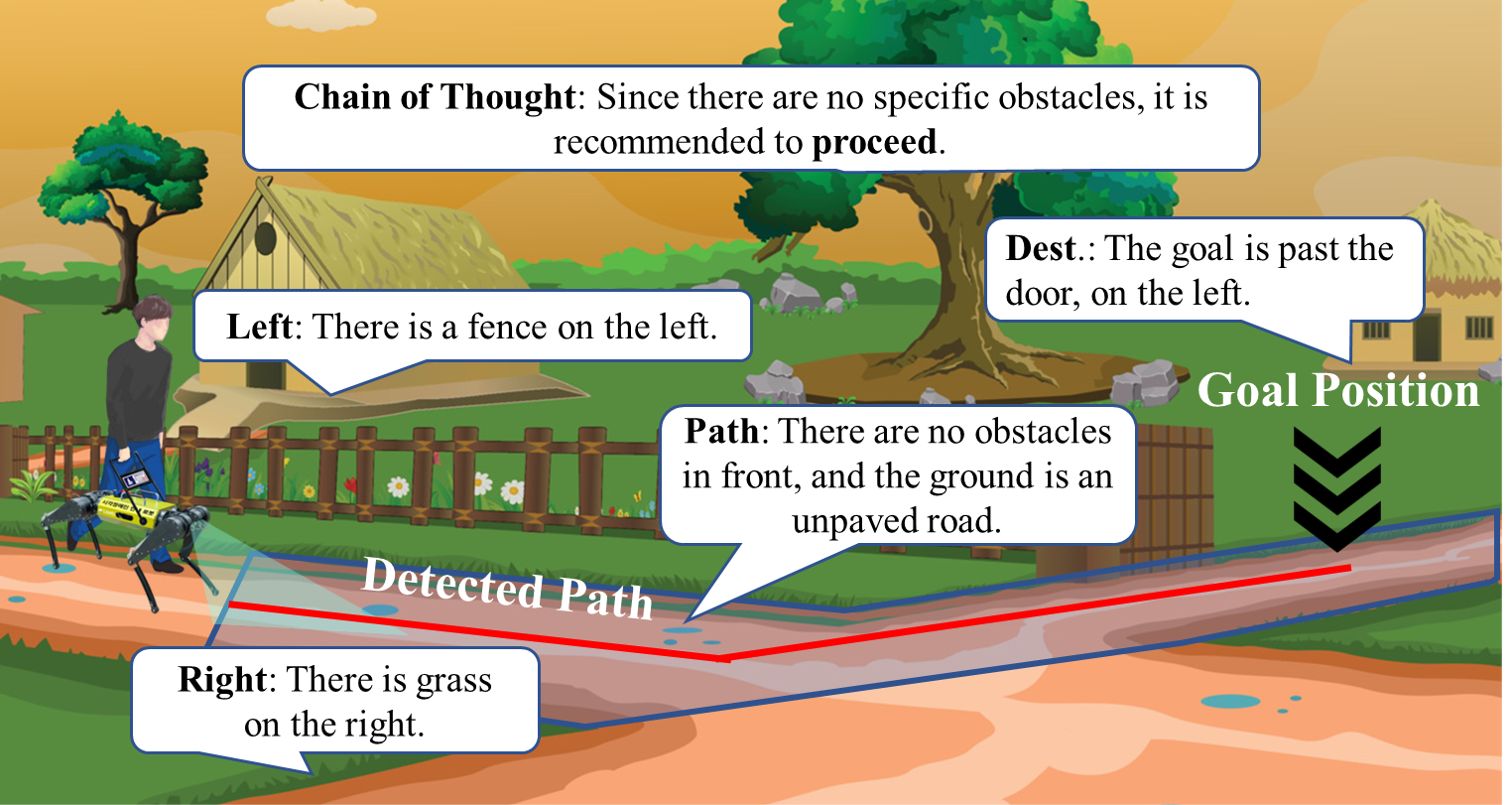} % adjust width as necessary
    \vspace{-15pt}
    \caption{Our space-aware instruction tuning method provides helpful and concise walking guidance to individuals with visual impairments.
   Given an image and a specified goal position as input, our method identifies a virtual path in 3D space and provides compact responses to five types of queries in a single inference: descriptions of the destination, the left and right sides of the path, the path itself, and a decision on whether the path is traversable, with a brief reason.    
   }
    \label{fig:intro}
    \vspace{-20pt}
\end{figure}

Although VLMs exhibit rich descriptive capabilities for images due to extensive pre-training on numerous images and a text corpus, they often struggle with interpreting spatial information such as spatial relationships and object positions \cite{chen2024spatialvlm}.
% This limitation restricts their ability to effectively understand and utilize these spatial aspects within the visual data \cite{Mert2023}\cite{Shengbang2024}.
The study \cite{Mert2023} has shown that VLMs frequently fail to accurately represent simple spatial relationships such as ``to the right of'' or ``behind'' in images, due to limitations in the CLIP vision encoder. 
For instance, when VLMs asked to distinguish between ``on the right of'' and ``on the left of'' in an image where a dog is positioned to the right of a tree, VLMs often perform at chance level. 
In VQA tests, similar failures of VLMs were observed in handling orientation and direction, e.g., identifying whether a rabbit is facing right or left, and positional and relational context, e.g., identifying whether glasses are on the right or left of a slipper \cite{Shengbang2024}. 
Even high-performance models, such as Gemini \cite{googleGemini} and GPT-4 \cite{openai2024gpt4technicalreport}, faced these challenges.
The highest accuracy in orientation tasks was only approximately 25\%, while positional and relational context tasks achieved around 45\% accuracy.

In this paper, we present the Space-Aware Instruction Tuning (SAIT) dataset, designed to help guide dog robots provide precise and effective walking guidance to the visually impaired. 
We also propose a novel automatic data generation pipeline that is applicable to any images in a walking environment. 
To integrate spatial awareness into the dataset, the pipeline explicitly extracts a depth map and identifies a virtual path to the goal position in 3D space. 
By showing only the relevant regions to the VLMs, it addresses the common challenge of distinguishing directions in standard VLMs.
To evaluate the model's ability to convey concise and meaningful information to the visually impaired, we manually annotated and created the Space-Aware Benchmark (SA-Bench) with an evaluation protocol.

Our key contributions are as follows:

\begin{itemize}
   \item[1)] We release the SAIT dataset and SA-Bench, along with an evaluation protocol to assess how effectively a VLM provides walking guidance for the visually impaired.
   \item[2)] We propose a novel data generation pipeline for automatically creating data to facilitate the understanding of three-dimensional space.
   \item[3)] We demonstrated the efficacy through comparisons experiments between the space-aware instruction-tuned model and other state-of-the-art algorithms.
\end{itemize}

%% file: tex/method.tex
\section{Method} 
\vspace{-8pt}
\input{tex/method_motivation} 
\input{tex/method_benchmark} 
\input{tex/method_pipeline}
\input{tex/method_training}
\input{tex/method_eval}

%% file: tex/method_motivation.tex
\subsection{Motivation}  
% or Motivation

Previous studies have shown the limitations of guide dogs in providing perceptual intelligence for the visually impaired, as well as the key considerations for describing specific object in the environment to assist the visually impaired for navigation \cite{Hochul2024}\cite{Hoogsteen2022}.
The findings of these studies suggest that the preferred methods of description and the frequency of requests among visually impaired individuals vary based on factors like physical balance, residual vision, age, whether they are early or late blind, and the type of mobility aid used, such as a white cane or guide dog.
For example, early blind participants, who asked fewer questions, tended to focus on identifying a building through its address, whereas late blind participants prioritized locating it directly from their own perspective. 
Individuals using a white cane preferred guidance-related descriptions over collision-related ones \cite{Hoogsteen2022}.
Some guide dog handlers want to be informed about every small obstacle, while others prefer to bypass minor ones like low-rise toe-trip hazards \cite{Hochul2024}. 
Despite the variety of opinions, \textit{a common preference was to receive \textbf{essential} walking information in a \textbf{concise} format.}

In this context, we conducted interviews to determine what descriptions are essential for the visually impaired if walking situations can be captured through the camera of a guide dog robot. 
A total of three face-to-face, in-depth interviews were conducted with two visually impaired individuals and four instructors from guide dog training schools.
Based on these findings, we can summarize the essential information in walking situations as follows:
\begin{itemize}
    \item \textit{A brief description should cover the left and right sides of the path, the path itself including any obstacles, and whether it is possible to walk along it.}
    \item \textit{If it is possible to walk to the destination, no explanation is needed.}
    \item \textit{If walking to the destination is not possible, provide a brief reason of the situation.}
\end{itemize}

\subsection{Problem Definition}
From the interview results, we formulate the problem as determining how to provide essential and concise information to the visually impaired.
To clarify, we assume that an image $I$ is captured by an RGB camera mounted on the guide dog robot, with a goal position (GP) representing the user's intended destination in the image.
The GP is a point $\mathbf{p_{goal}} = (x, y)$ in the image $I$. 
The first stage is determining the path to the GP.
At this stage, we set the path width to 2 meters, incorporating a safety buffer with the larger value within the range of 1.2 to 1.5 meters, which is known to accommodate the safe movement of a visually impaired person and a guide dog \cite{MacLennan2015}.
Based on the detected path, descriptions are given for the path, its left and right sides, and the destination, represented as $t_{path}$, $t_{left}$, $t_{right}$, and $t_{dest}$, respectively.
Since the descriptions must be concise and essential, we focus on nearby objects or obstacles, as a survey \cite{Hoogsteen2022} found they are more useful than distant ones.
After providing these descriptions, a decision $t_{reco}$ is made on whether it is possible to proceed along the path, with a brief explanation of the reasons for the decision.
Thus, by receiving an image $I$ and a query prompt $q_{goal}$ containing a goal position $\mathbf{p_{goal}}$ as input, a VLM, $f$, provides four types of spatial descriptions and makes decisions based on that information as follows:
\begin{equation}
t_{dest}, t_{left}, t_{right}, t_{path}, t_{reco} = f(I, q_{goal})
\end{equation}

%% file: tex/method_benchmark.tex
\subsection{Dataset Collection and Benchmark}
\label{sec:dataset_collection_and_benchmark}

\begin{table*}[t]
%\vspace{-8mm}
\caption{Summary of datasets}
%: the basis of the SAIT dataset and SA-Bench, focusing on landmarks and obstacles in the walking environment.
\setlength{\tabcolsep}{5pt}
\vspace{-3mm}
\centering
\begin{tabular}{|c|c|l|c|c|c|}
\hline
\textbf{Dataset} & \textbf{Purpose} & \textbf{Annotations} &\textbf{\#classes}& \textbf{\#images} & \textbf{\#instances} \\
\hline
\multirow{2}{*}{VIN}& Pedestrian guidance& Door, Elevator, Escalator, Stairs, Pedestrian traffic light,&\multirow{2}{*}{7}&\multirow{2}{*}{7,493}&\multirow{2}{*}{25,480} \\
&landmark detection& Entrance of subway station, Subway ticket gate && & \\
\cline{1-6}
\multirow{5}{*}{SideGuide\cite{Kibaek2020}}&{Moving object}&Person, Stroller, Car, Wheelchair, Bus, Dog, Truck, &\multirow{5}{*}{29}&\multirow{5}{*}{349,741}&\multirow{5}{*}{3,367,063}\\
&detection& Cat, Bicycle, Carrier, Motorcycle, Movable signage, Scooter &&&\\
\cline{2-3}
& \multirow{2}{*}{Fixed object}& Tree trunk, Bus/Taxi stop, Potted plant, Kiosk, Traffic light,  &&&\\
&\multirow{2}{*}{detection}&Fire hydrant, Traffic sign, Parking meter, Pole, Bollard, Bench,  &&&\\
&&Barricade, Chair, Power controller, Table, Traffic light controller&&&\\
\hline
SAIT&Instruction-tuning&Goal position, Queries, Descriptions, Object classes \& locations, Path array&-&20,000&-\\
\hline
SA-Bench&Evaluation&Goal position, Descriptions, Path array, Passibility (go/stop)&-&1,000&-\\
\hline
\end{tabular}
\label{tab:dataset}
\vspace{-18pt}
\end{table*}
%\color{red}{with reason}

\vspace{-3pt}
\subsubsection{Visually Impaired Navigation (VIN) dataset}
We collected image data in a scenario reflecting everyday life where visually impaired individuals move from their homes to a destination using the subway as a representative means of public transportation.
Specifically, we captured 7,493 images along 75 distinct routes—starting at random apartment buildings, passing through subway stations, and ending at various destinations like the city hall civil affairs office—using an RGB camera positioned approximately 80 cm above the ground to match the height of the guide dog robot.
%Considering various times of day, weather, and seasonal changes, we collected data over six months.
In particular, crosswalks with pedestrian traffic lights, where visually impaired individuals may experience the most difficulty, were included in the routes to capture crossing street situations.
We defined essential pedestrian guidance landmarks for the visually impaired, captured images of them from various angles along planned routes, and annotated their bounding boxes as described in Table \ref{tab:dataset}.

\subsubsection{SideGuide dataset} Furthermore, to incorporate various moving and fixed obstacles in a wider variety of walking environments, we used a total 349,741 images in the pedestrian walking dataset with 3,367,063 object instances \cite{Kibaek2020} as described in Table \ref{tab:dataset}.

\subsubsection{Space-Aware Instruction Tuning (SAIT) dataset}
% 데이터를 생성해서 학습하려는 동기
% VLMs are known to have limited spatial understanding and reasoning abilities \cite{Mert2023}\cite{Shengbang2024}.
% If VLMs are used to explain the current situation to visually impaired individuals based on images, there is a risk of incorrect descriptions, such as confusing left and right or misunderstanding the intended path or direction. Therefore, if a dataset with accurate spatial descriptions were available, it could be used to train VLMs to develop better spatial understanding.
% However, 
To the best of our knowledge, there are currently limited image datasets captured from the pedestrian's viewpoint \cite{Kibaek2020}\cite{Karnan2022}\cite{Zhao2023}, and no dataset with precise spatial descriptions.
To address this, we composed a novel dataset with images captured from the pedestrian's viewpoint and their spatial descriptions to provide walking guidance to the visually impaired.
We sampled 20,000 images from the SideGuide to construct our SAIT dataset.
The dataset, designed to enable the VLM to understand images in a space-aware manner, contains object-centered descriptions of the destination, path, and left/right sides of the path, all based on a virtual path to a GP. 
It also provides walkability information based on these descriptions.
Landmark and obstacle data were provided by separate object detectors to improve dataset accuracy, and details are described in Section \ref{sec:pipeline}.
% Leveraging the obstacle information and images, we created an instruction-tuning dataset for training VLMs based on the auto-generation pipeline described in Section \ref{sec:pipeline}.

\subsubsection{Space-aware benchmark (SA-Bench)}
To evaluate the accuracy and conciseness of the guiding sentences generated by the instruction-tuned VLM, we created a benchmark set by extracting images from VIN and SideGuide that are mutually exclusive from the SAIT dataset.
The benchmark set consists of 1,000 images, including 500 `go' scenarios and 500 `stop' scenarios, determined based on the decision of whether the path to the GP is passable.
For example, even if there are no obstacles to a GP in an image, it is classified as a `stop' if the pedestrian traffic light is red.
For each image, five types of text descriptions describing the walking situation with a GP and path to it were manually annotated as described in Table \ref{tab:benchmark}.

\begin{table}[b]
\vspace{-5mm}
\caption{Annotations of the SA-Bench}
\vspace{-3mm}
\centering
\begin{tabular}{|c|c|l|}
\hline
\textbf{Category}&\textbf{Format}&\textbf{Content}\\
\hline
GP&(x, y)&Human-assigned position in an image\\
% 이미지 내의 갈 법한 곳을 랜덤하게 사람이 지정
\hline
\multirow{2}{*}{Path Array}&Array of&\multirow{2}{*}{Path to the GP}\\
&(x, y)&\\
% Path Array&Array of (x, y)&Path to the GP\\
%&(x, y)&\\
\hline
Dest.&Text&Description of the destination\\
\hline
Left&Text&Description of the left side of the path\\
\hline
Right&Text&Description of the right side of the path\\
\hline
Path&Text&Description of the path itself\\
\hline
\multirow{2}{*}{Reco.}&\multirow{2}{*}{Text}&Decision on whether the path is passable,\\
&& with a simple reason\\
\hline
\end{tabular}
\label{tab:benchmark}
\end{table}

%% file: tex/method_pipeline.tex
\vspace{-8pt}
\subsection{Training Dataset Generation Pipeline}
\label{sec:pipeline}
\vspace{-2pt}

\begin{figure*}[ht]
    \centering
%    \vspace{-8mm}    
    \includegraphics[width=\linewidth]{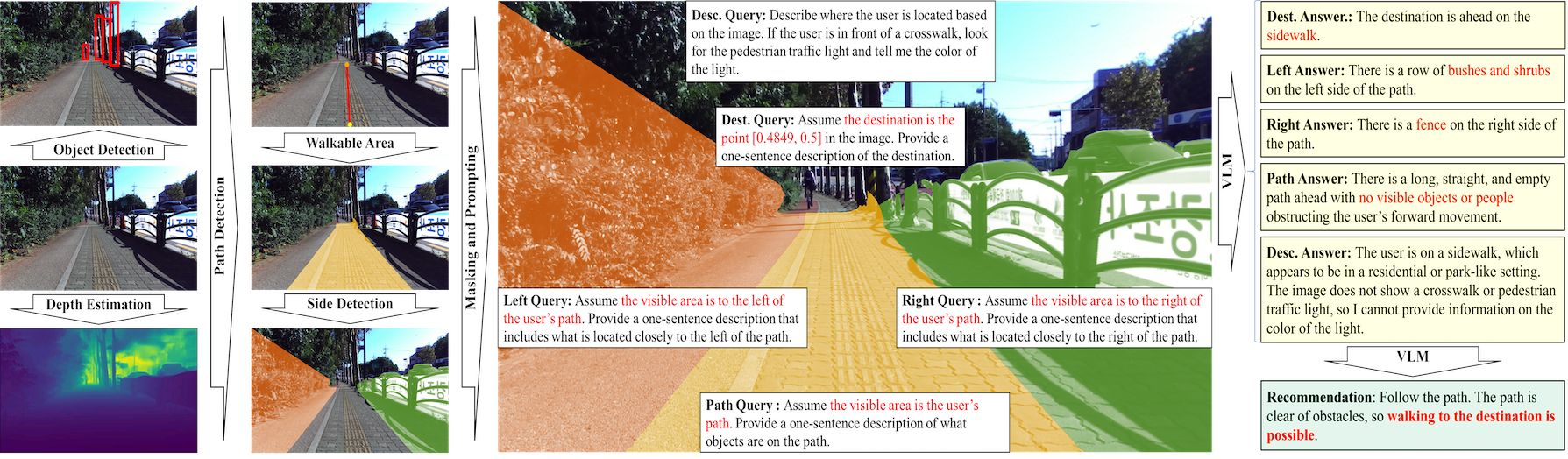} % adjust width as necessary
    \vspace{-8mm}
    \caption{Our automatic dataset generation pipeline that focuses on the virtual path to the destination in 3D space and the surroundings. Given an input image and a goal position, we extract object locations and generate a depth map to identify walkable areas and surrounding environments, which are then utilized as inputs for a VLM. 
    Subsequently, we generate five distinct sentences through separate queries and integrate them to produce a recommendation statement on whether visually impaired individuals can traverse the specified path with a brief reason. }
    \label{fig:pipeline_flow}
    \vspace{-6mm}
\end{figure*}

% 자동 생성 과정에 대한 전체적인 구조
Overall process of our automatic dataset generation pipeline is presented in Fig.~\ref{fig:pipeline_flow}.

\subsubsection{Acquiring object and depth information}
\label{sec:object_detection}

To provide landmark information, as well as the locations of moving and stationary obstacles within the images, we trained YOLOv8 models \cite{Jocher_Ultralytics_YOLO_2023} on each of the two datasets, VIN and SideGuide. 
Due to the imbalance in the quantity of these two datasets and the exclusive labels of objects, we trained separate detectors. For each input image, we obtain 7 types of landmarks and 29 types of obstacles from the two object detectors, which are then added in input prompts for the VLM.
The distance information was estimated using Depth-Anything-V2, a monocular depth estimation algorithm \cite{depth_anything_v2}. 

\subsubsection{Goal setting and pathfinding from viewer position to destination}
\label{sec:GP_and_PP}
To describe an image based on the path, both the GP and the path to the GP must be provided. The reason for specifying the GP is that even with the same image, different GPs and paths result in different descriptions. 
In the Guide Dog Robot system, the GP is determined through global path planning. However, for generating training data, this approach is not feasible, so we randomly generated GPs in plausible locations within the image. The generation method assumes the bottom center of the RGB image as the starting point of an user, then randomly sets a direction within a 45-degree angle to the left or right, and finally places the GP 10 meters away from the starting point.
The path from the start point to the GP was assumed to be a straight line. 

\subsubsection{Masking and prompting VLM for descriptive information}
Since VLMs are known to understand coordinates \cite{liu2023llava}, the simplest way to obtain a description of the path to the GP is to provide an image along with a related prompt (e.g., ``Describe the path to the coordinates (x, y)''. However, we found that VLMs are unable to estimate a path based solely on GP coordinates, and they struggle with understanding the path and its left and right sides \cite{Mert2023}\cite{Shengbang2024}.
To address this challenge, instead of providing a single prompt describing the destination, the left and right sides of the path, and the path itself all at once, we input separate prompts for each area along with images showing only those specific sections, as illustrated in Fig.~\ref{fig:pipeline_flow}. 
To provide only the relevant regions of the image, we estimated the path area and masked either the left or right side of the path. This approach helped reduce confusion for the VLM when describing each section and improved accuracy. By splitting the prompts, we were able to overcome the issue of degraded performance in VLMs when handling long prompts \cite{Han2024lminfinite}\cite{levy2024tasktokensimpactinput}. 

Specifically, once the GP is determined, the path is calculated to have a width of 2 meters, centered on the ground, using the camera's intrinsic parameters and the depth values obtained in Section~\ref{sec:object_detection}. However, due to the inaccuracies in both the depth values and the camera's intrinsic parameters, corrections were made using the actual sizes of objects such as pavement blocks, cars, and people. The formula to calculate the left and right points of the path, $\mathbf{P_{i, left}, P_{i, right}}$, using the $i$-th point on the path at image coordinates $\mathbf{P_{i, path}} = (x_{i, path}, y_{i, path})$, depth value $z_i$, and the camera's focal length $\ell_x$ and $\ell_y$ is as follows.
\begin{equation}
% \vspace{-20pt}
\mathbf{p}_\text{i, left} = \begin{bmatrix} x_i -  \frac{2}{z_i}\ell_x \\ y_i \end{bmatrix}, \quad \mathbf{p}_\text{i, right} = \begin{bmatrix} x_i +  \frac{2}{z_i}\ell_x \\ y_i \end{bmatrix}
\end{equation}
The left area of the path is shown by displaying only the region formed by connecting the top-left corner, bottom-left corner, and the left points of the path, $\mathbf{p}_\text{i, left}$ of the image. Similarly, the right area is displayed by connecting the top-right corner, bottom-right corner, and the right points of the path, $\mathbf{p}_\text{i, right}$ to form a polygon. The path area is displayed by sequentially connecting the left points $\mathbf{p}_\text{i, left}$ and the right points $\mathbf{p}_\text{i, right}$. 
In these displayed areas, areas located a certain distance behind the destination are removed to exclude irrelevant background elements, such as the sky. 
% that are not necessary for the description.
% An example of these results is shown in Fig.~\ref{fig:pipeline_flow}. 
For the destination, since the VLM can understand coordinates, the original image is provided without masking. 
Different prompts are used for each area’s description as described in Fig. \ref{fig:pipeline_flow}.
The prompts include the class and location 
of the detected objects in Section~\ref{sec:object_detection}. Only the regions of objects that overlap with the non-masked area are provided.

\subsubsection{Path navigability decision using VLM descriptions}
Based on the path-centered descriptions obtained earlier, we queried the VLM to determine whether the path was walkable. Prior to this query, a full image description focused on crosswalks and traffic lights had already been included in the previous descriptions. Additionally, we incorporated rules related to safety (e.g., if the user is in front of a crosswalk with a red pedestrian signal or in a hazardous area, they should stop). 
%The same VLM, LLaVA, was used for this inference.
To summarize this process, the input for the VLM, $f$, for each image description $t_i$, can be formulated using the image $I$, mask $M$, object information $k$, query prompt $q$, and the concatenation operator `;'. In particular, the query prompt $q_{\text{dest}}$ contains the goal position $\mathbf{p}_{\text{goal}}$.
\begin{align}
\begin{split}
    I_i &= I \cap M_i, \quad k_i = k \cap M_i \\
    t_i &= f(I_i, (q_i ; k_i)), \quad i \in \{\text{left}, \text{right}, \text{path}\} \\
    t_i &= f(I, (q_i ; k)), \quad i \in \{\text{dest}, \text{desc}\} \\
    t_i &= f(I, (t_{\text{dest}} ; t_{\text{left}} ; t_{\text{right}} ; t_{\text{path}} ; t_{\text{desc}} ; q_i)),  i = \text{reco}
\end{split}
\end{align}

%% file: tex/method_training.tex
\begin{figure*}[t]
  % \vspace{-8mm}
  \centering
  \includegraphics[width=\linewidth]{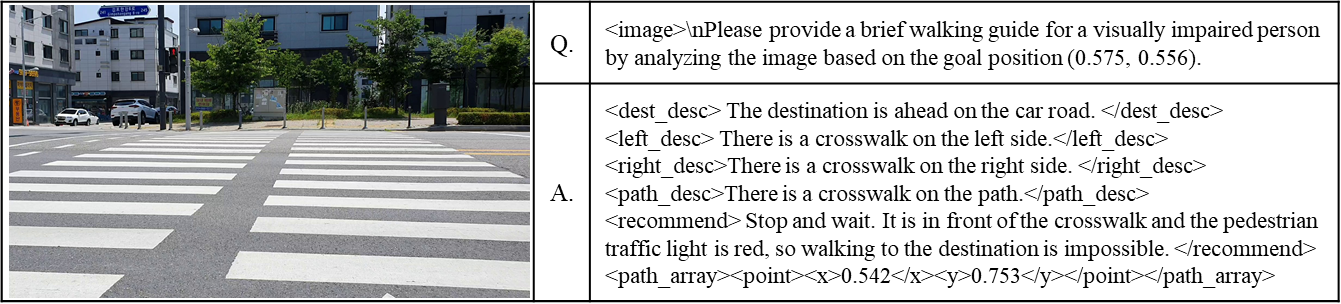}
    \vspace{-8mm}
    \caption{An image and annotations of the SAIT dataset: queries were generated based on the goal positions, and answers were structured in XML format. Note that words like \texttt{<dest\_desc>} are plain text in the answer, with only the image token in the query treated as a special token.}
  \label{fig:conv_template}
  \vspace{-5mm}
\end{figure*}
\subsection{Training} 
\vspace{-2pt}
Utilizing constructed the SAIT dataset, the automatically generated walking guidance annotations were transformed into a conversation format for VLM training.
As illustrated in Fig. \ref{fig:conv_template}, the query includes the image token and the GP, while the answer contains five types of descriptions and the path array to the GP in XML format.
In other words, given a query containing a GP, our instruction-tuned Space-Aware-VLM (SA-VLM) can predict five descriptions and a path to the GP in a single inference.
Note that the words such as \texttt{<dest\_desc>} are not special tokens but plain text.

%% file: tex/method_eval.tex
\vspace{-5pt}
\subsection{Evaluation Criteria}
\label{sec:eval}
\subsubsection{LLM Judge}
% LLM-as-a-judge
LLM Judge (referred to as LLM-as-a-judge in \cite{zheng2023llmjudging}) is a proposed method for evaluating LLM responses to open-ended questions, involving the use of structured prompts with strong LLMs \cite{zheng2023llmjudging}. 
% 뭔지 무엇을 찾아내고, 어떻게 측정하는지
With a strong LLM as a judge, the prompt is structured to compare the reference answer with the assistant's answer, identifies any errors, corrects them, and then assigns a score between 1 and 10 based on this comparison.
This approach achieved a rating agreement level equivalent to human evaluations.
In our experiment, we adapted the prompt for reference-guided multi-turn single-answer grading from \cite{zheng2023llmjudging} by modifying it to a single-turn format and replacing conversation-related terms with description-related ones.
% The average number of words used in the response was also included as a metric to assess the conciseness of the answers.

\subsubsection{Text summarization metrics: }
%METEOR, BERTScore, ROUGE
To assess the quality of the generated text, we also utilize three widely-used metrics: METEOR\cite{banerjee2005meteor}, ROUGE\cite{lin2004rouge}, and BERTScore\cite{zhang2019bertscore}. These metrics capture different dimensions of similarity between the generated outputs and reference texts, ensuring a thorough evaluation.

%% file: tex/results.tex
\section{Experiments}
In this section, we conduct comparative experiments to demonstrate the impact of our SAIT dataset on SA-Bench described in \ref{sec:dataset_collection_and_benchmark}. 
We also perform experiments to validate the effectiveness of the pipeline described in Section \ref{sec:pipeline}.

\vspace{-5pt}
\subsection{Experiment Details}
For our SA-VLM, we selected LLaVA-OneVision \cite{li2024llavaonevisioneasyvisualtask}  as our base model, a VL multimodal model to improve comprehension in complex visual scenarios.
We fine-tuned the 7B model across all components: the vision encoder, MLP adapter, and the large language model on our SAIT dataset with a batch size of 64 for one epoch. 
The learning rate was set to 1e-5 with a cosine scheduler using 8 NVIDIA RTX 6000 Ada GPUs for around 4 hours.
For our automatic generation pipeline, we used LLaVA-v1.6-34b \cite{liu2023improved} without additional fine-tuning. 
Including object detection, depth estimation, masking, and generating descriptions, the process took about 62 seconds per image, using 4 NVIDIA RTX 6000 Ada and 4 Quadro RTX 6000 GPUs.

\input{tex/results_comparison}

\input{tex/results_pipeline}

%% file: tex/results_comparison.tex
\subsection{Comparison Experiments on SA-Bench}
\label{sec:comp_sota}
We conducted comparative experiments with our SA-VLM and a variety of VLMs of different sizes, including 7B, 13B, 34B, 72B, as well as the widely-used GPT-4o.
For GPT-4o, we utilized specific versions: gpt-4o-mini-2024-07-18 and gpt-4o-2024-08-06.
To ensure fair comparisons, we adopted the multi-turn query approach for methods other than SA-VLM because it demonstrated higher performance than the single-turn query approach, as in Section \ref{sec:single_vs_multi}.
Specifically, we first generated four types of sentences using in-context learning.
As required by SA-Bench in Table \ref{tab:benchmark}, these sentences describe the GP area (Dest.), the path to the GP (Path), and the left and right sides of the path (Left, Right), with four separate inferences.
Then, a full image description is generated once more. 
The five collected sentences are then used as input prompts for them.
Based on these prompts, a recommendation (Reco.) sentence is produced through a chain of thought (CoT) process to determine whether it is possible to reach the GP.
In short, our SA-VLM performs a single inference, while the other approaches require a total of six inferences to generate the five sentences required by SA-Bench. 
For our evaluation metric, we used the LLM Judge based on GPT-4o as explained in Section \ref{sec:eval}.

\setlength{\tabcolsep}{1pt}
\begin{table}[t!]
\caption{Performance comparisons on SA-Bench \\with SOTA algorithms using LLM-judge. Dest.: Destination, Reco.: Recommendation, Inf: Inference}
\vspace{-3mm}
\centering
\begin{tabular}{|l|c|c|c|c|c||c||c|c|}
\hline
\multirow{3}{*}{\textbf{Models}} & \multirow{2}{*}{\textbf{Dest.}} & \multirow{2}{*}{\textbf{Left}}& \multirow{2}{*}{\textbf{Right}} & \multirow{2}{*}{\textbf{Path}} & \multirow{2}{*}{\textbf{Reco.}} & \multirow{2}{*}{\textbf{Avg.}} & \textbf{\# of} & \textbf{Inf.} \\
&&&&&&&\textbf{Words} & \textbf{Time}\\
\cline{2-9}
& \multicolumn{6}{c||}{\(\uparrow\) Higher is better} & \multicolumn{2}{c|}{\(\downarrow\)} \\
\hline
\hline
llava-v1.6-vicuna-7b\cite{liu2023improved} & 5.14 & 2.64 & 2.81 & 1.85 & 3.57 & 3.20 &12.75 & 11.66\\ \hline
llava-v1.6-vicuna-13b\cite{liu2023improved} & \textbf{5.25} & 1.44 & 1.22 & 2.11 & 4.07 & 2.82 &14.12 & 15.80 \\ \hline
llava-v1.6-34b\cite{liu2023improved} & 4.49 & 3.03 & 2.99 & 2.43 & 4.70 &3.53& 12.56 & 51.89 \\ \hline
llava-next-72b\cite{li2024llavanext-strong} & 4.57 & 2.44 & 2.46 & 3.41 & 4.90 &3.55& 11.96 & 46.63 \\ \hline
gpt-4o-mini\cite{openai2024gpt4technicalreport} & 3.32 & 2.56 & 2.47 & 2.83 & 4.31&3.10 & 14.29 & 34.66 \\ \hline
gpt-4o\cite{openai2024gpt4technicalreport} & 3.51 & 3.14 & 3.19 & \textbf{3.66} & 4.39 & 3.58&12.11 & 40.86 \\ \hline
\hline
\textbf{SA-VLM-7b (ours)} & 4.43 & \textbf{3.36} & \textbf{3.25} & 2.64 & \textbf{4.93}&\textbf{3.72} & \textbf{11.95} & \textbf{6.81} \\ 
\hline
\end{tabular}
\label{tab:comparison}
\vspace{-8pt}
\end{table}

\setlength{\tabcolsep}{2pt}
\begin{table}[t!]
% \vspace{-10pt}
\caption{Performance comparison on SA-Bench \\with the base model using LLM-judge}
\vspace{-3mm}
\label{tab:with_train_performance}
\centering
\begin{tabular}{|l|c|c|c|c|c|c|c|}
\hline
\multirow{2}{*}{\textbf{Models}} & \multirow{2}{*}{\textbf{Dest.}} & \multirow{2}{*}{\textbf{Left}} & \multirow{2}{*}{\textbf{Right}} & \multirow{2}{*}{\textbf{Path}} & \multirow{2}{*}{\textbf{Reco.}}& \textbf{\# of} & \textbf{Inf.} \\
&&&&&&\textbf{Words}&\textbf{Time}\\
\hline
\hline
llava-ov-qwen2-7b-si \cite{li2024llavaonevisioneasyvisualtask} & 3.55 & 2.43 & 2.29 & 2.25 & 4.05 & 14.21 & 20.39\\ 
\hline
\textbf{SA-VLM-7b (ours)} & \textbf{4.43} & \textbf{3.36} & \textbf{3.25} & \textbf{2.64} & \textbf{4.93} & \textbf{11.95} & \textbf{6.81} \\ 
\hline
\end{tabular}
\vspace{-20pt}
\end{table}

As shown in Table \ref{tab:comparison}, our SA-VLM demonstrated superior average performance across five types of responses, even surpassing larger models such as GPT-4o.
This is because general VLMs have not yet effectively learned space-related concepts, making it difficult to distinguish between left and right. 
Thus, they exhibited lower performance in responses of the Left and Right type.
This weakness also impacts CoT reasoning when generating the Reco. sentence.
Interestingly, smaller models such as LLaVA-v1.6-Vicuna 7B and 13B performed better in the Dest. type compared to larger models but exhibited lower performance in other types. 
This discrepancy can be attributed to the nature of SA-Bench, which naturally includes walking environments with a relatively large number of sidewalks.
Smaller models tend to generate biased responses or select the first example from the query prompt, such as `The destination is ahead on the sidewalk,' without adequately analyzing the image. This behavior artificially inflated the score for the Dest. type, while reducing accuracy in other types.

The model shows the best performance in terms of average word count and inference time across five sentence types.
This means that visually impaired individuals can receive walking guidance that is more concise and faster, allowing them to navigate more efficiently without unnecessary detail or complexity.
Furthermore, we also measured performance of our base model, LLaVA-OneVision-7b.
As shown in Table \ref{tab:with_train_performance}, the results clearly demonstrate that training was effective, improving performance across all aspects.

\begin{figure}[b!]
    \centering
    \vspace{-15pt}
    \includegraphics[width=\linewidth]{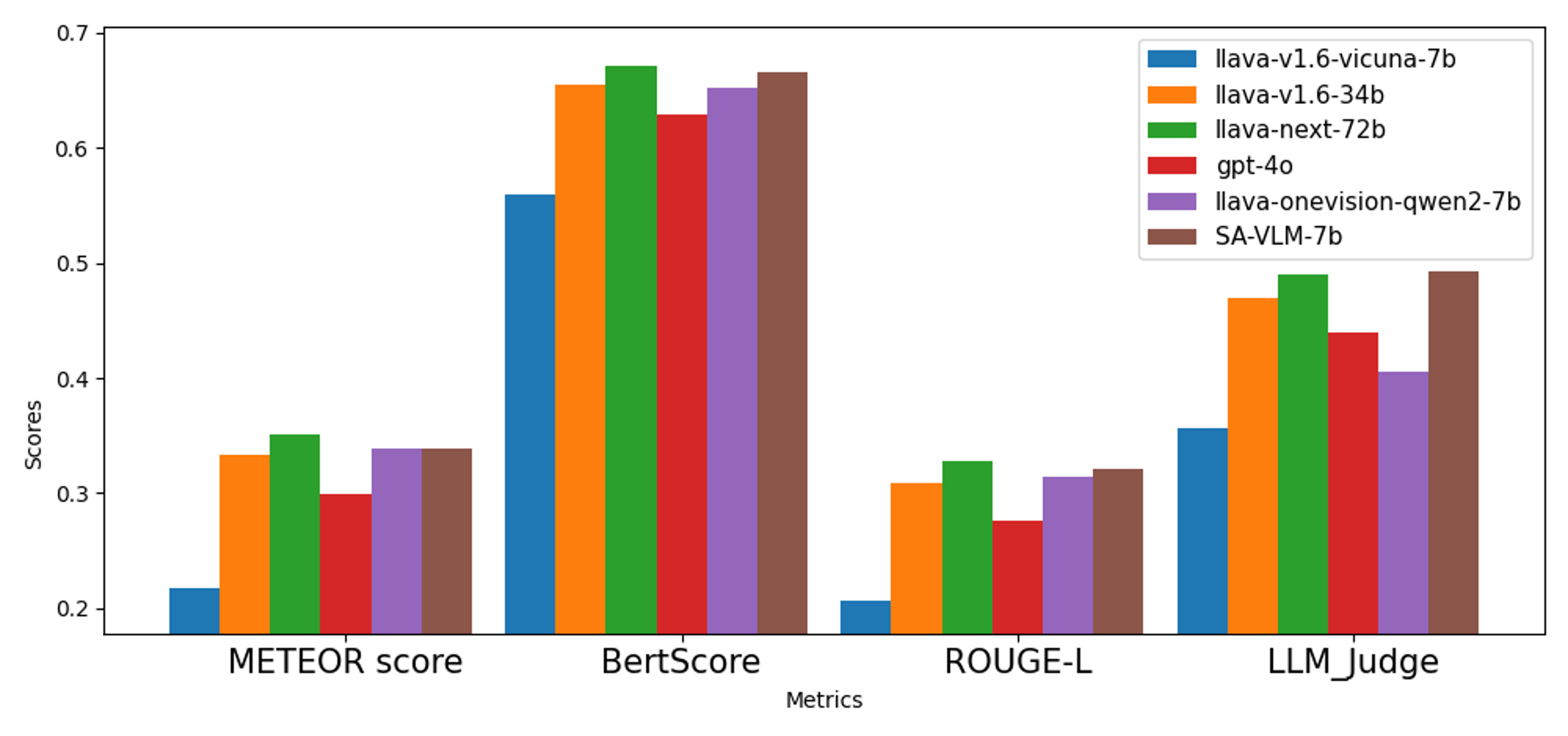} % adjust width as necessary
    \vspace{-25pt}
    \caption{Comparisons of algorithms across different metrics for the generated `Reco.' sentences. We chose LLM Judge as our main metric because it provides scoring based on the contextual meaning of the generated sentences.}
    \label{fig:metric_comp}
    % \vspace{-5mm}
\end{figure}

Lastly, we evaluated the key approaches using METEOR, BERTScore, and ROUGE-L metrics.
As shown in Fig. \ref{fig:metric_comp}, both llava-next-72b and our model achieved the best performance across all metrics. However, ROUGE-L and METEOR were not fully suitable as evaluation metrics for SA-Bench, since they prioritize word overlap rather than the semantic meaning of sentences. 
BERTScore also exhibited an issue, where it rated semantically opposite words like ``go'' and ``stop'' similarly. 
This error arises because these words are more related to navigation, leading to high embedding similarity and inflated scores.
In contrast, LLM Judge provided more reasonable evaluations by assessing sentences based on their contextual meaning, offering a more accurate reflection of performance in SA-Bench.

%% file: tex/results_pipeline.tex
\subsection{Pipeline Evaluation}
We evaluated effectiveness of the pipeline by applying it on SA-Bench. The GP is provided by the SA-Bench, and the resulting descriptions were compared with the ground truth.

\subsubsection{Single-turn query vs Multi-turn query}
\label{sec:single_vs_multi}
When querying a VLM for scene descriptions, one can easily consider structuring the desired information into a single-turn query, where all relevant details are asked in one prompt.
We compared the single-turn query approach with a multi-turn query method, used in our pipeline, in terms of performance and speed.
In our experiment, we structured the single-turn query prompt so that numbered questions for each region and final recommendation were input into VLMs, and the answers were returned in a corresponding numbered format.

As shown in Table~\ref{table:one_turn}, the multi-turn query approach resulted in more accurate descriptions across four types compared to the single-turn method except for the Dest. type. 
This pattern of the single-turn query results mirrors that of the smaller model in Table~\ref{tab:comparison}, where the model's reasoning abilities declined. 
Upon analyzing the results, we found that the single-turn query often selected answers from examples in the query prompt rather than analyzing the image.
This aligns with previous findings that longer prompts in single-turn query can weaken reasoning abilities, as highlighted in prior studies \cite{Han2024lminfinite}\cite{levy2024tasktokensimpactinput}. 
In terms of processing speed and description length, the single-turn method performed better, but the difference was not significant. 
This is because, in VLMs, response time scales with the length of the input and output text, and there was no significant difference in the total number of output words between single-turn and multi-turn queries.

\begin{table}[!ht]
    \vspace{-10pt}
    \caption{Performance comparisons on SA-Bench with single-turn query and multi-turn query using LLM Judge}
    \vspace{-7pt}
    \centering
    \begin{tabular}{|l|c|c|c|c|c||c|c|}
    \hline
        \textbf{Methods}         & \textbf{Dest.}  & \textbf{Left} & \textbf{Right} & \textbf{Path} & \textbf{Reco.} & \textbf{\# Words} & \textbf{Inf. Time} \\ \hline\hline
        Multi-turn     & 4.49 & \textbf{3.03} & \textbf{2.99} & \textbf{2.43} & \textbf{3.53} & 12.56 & 51.89 \\ \hline
        Single-turn    & \textbf{4.91} & 1.46 & 1.45 & 1.71 & 2.72 & \textbf{10.54} & \textbf{34.67} \\ \hline
    \end{tabular}
    \label{table:one_turn}
    \vspace{-8pt}
\end{table}

\subsubsection{Masking region vs Prompting region} 
This experiment examines whether masking the image to a specific spatial region helps the VLM better understand the target of the description. 
The targets for description include the destination represented by coordinates, the path represented by the starting and goal points, and nearby objects located to the left and right of the path.
In the region-masked method, only the area corresponding to the target is visible, with the rest of the image masked, while a standard prompt is provided to the VLMs.
In the region-prompted method, the target area information, such as coordinates, is included directly in the prompt, and the entire image is presented to the VLMs.
Table~\ref{table:masked_prompted} presents the performance of each target area for the two methods on the LLM Judge and the difference between the two methods was pronounced.

\begin{table}[!ht]
    \vspace{-8pt}
    \caption{Performance comparisons on SA-Bench with masked-image and region-prompt methods using LLM Judge}
    \centering
    \vspace{-3mm}
    \begin{tabular}{|l|c|c|c|c|}
    \hline
        \textbf{Methods} &\textbf{ Dest.}  & \textbf{Left} & \textbf{Right} & \textbf{Path} \\ \hline\hline
        Masked-Image    & 3.53 & \textbf{3.26} & \textbf{3.20} & \textbf{2.93} \\ \hline
        Region-Prompt  & \textbf{4.52} & 3.09 & 2.88 & 2.29 \\ \hline
        %Masked-Image (BERTScore)  & 0.6803 & \textbf{0.7680} & \textbf{0.7675} & 0.6821 \\ \hline
        %Region-Prompt (BERTScore)& \textbf{0.7399} & 0.7674 & 0.7636 & \textbf{0.6857} \\ \hline
    \end{tabular}
    \label{table:masked_prompted}
    \vspace{-8pt}
\end{table}

For destination descriptions, the region-prompt method using the full image and coordinates information in the prompt improved performance. On the other hand, the masked-region method using masked images with ordinary prompt enhanced the VLM's ability to describe the left, right, and path areas. This is because VLMs and GPT-based models are known to understand and match coordinates with images, making it possible to generate accurate descriptions based on coordinates alone \cite{liu2023llava}. However, since the models lack the ability to set paths and describe the left, right, and path areas based on those paths, it was found to be more effective to use masking to show only the relevant areas and obtain descriptions for them.

%% file: tex/conclusion.tex
\section{Conclusion}
In summary, we introduced a space-aware instruction tuning method aimed at improving guide dog robots' ability to provide accurate and effective walking guidance to the visually impaired.
We developed a novel space-aware data generation pipeline and used it to create the SAIT dataset. 
Additionally, we proposed SA-Bench and an evaluation protocol to enable the assessment of spatial awareness in VLMs.
Through comprehensive comparisons with state-of-the-art algorithms, we demonstrated the effectiveness of our approach, highlighting its potential in enhancing spatial guidance capabilities.

However, relying solely on an automated data generation pipeline can inadvertently lead to biased or noisy descriptions, resulting in lower LLM judge scores for SA-VLM. 
Hence, incorporating human-generated data is necessary to correct these inaccuracies and ensure a more robust SA-VLM.
Another challenge lies in the lack of real-world experiments, which impedes our understanding of how the method would perform under a real-world conditions such as inference latency or variations in weather and illumination. 
Future work should therefore focus on both enhancing data quality and conducting thorough real-world validations—including user studies with visually impaired individuals—to fully ascertain the method’s practical applicability.